%% file: main.tex
\pdfoutput=1

\documentclass[11pt]{article}

\usepackage{EACL2023}

\usepackage{times}
\usepackage{latexsym}

\usepackage[T1]{fontenc}

\usepackage[utf8]{inputenc}

\usepackage{graphicx}
\usepackage{caption}
\usepackage{microtype}
\usepackage{amsmath}
\usepackage[linesnumbered, ruled]{algorithm2e}

\SetCommentSty{mycommfont}
\usepackage{setspace}

\usepackage{booktabs}
\title{Focusing on Context is NICE:\\Improving Overshadowed Entity Disambiguation}

\author{Vera Provatorova$^1$, Simone Tedeschi $^{2,3}$, Svitlana Vakulenko$^4$\\ {\bf         Roberto Navigli}$^2$ \normalfont{and} {\bf Evangelos Kanoulas}$^1$\\
$^1$University of Amsterdam, $^2$Sapienza University of Rome\\
$^3$Babelscape, Italy, 
$^4$Amazon Alexa AI\\
\texttt{\{v.provatorova, e.kanoulas\}@uva.nl} \\
\texttt{\{tedeschi, navigli\}@diag.uniroma1.it} \\
\texttt{svitlana.vakulenko@gmail.com }}

\begin{document}
\maketitle
\begin{abstract}
Entity disambiguation (ED) is the task of mapping an ambiguous entity mention to the corresponding entry in a structured knowledge base. Previous research showed that entity overshadowing is a significant challenge for existing ED models: when presented with an ambiguous entity mention, the models are much more likely to rank a more frequent yet less contextually relevant entity at the top. Here, we present NICE, an iterative approach that uses entity type information to leverage context and avoid over-relying on the frequency-based prior. Our experiments show that NICE achieves the best performance results on the overshadowed entities while still performing competitively on the frequent entities. 
\end{abstract}
\input{1_introduction}
\input{2_methods}
\input{3_experiments}

\input{4_big_table}
\input{4_results}
\input{5_related_work}
\input{6_conclusion}
\input{7_limitations}

\bibliography{anthology, references}
\bibliographystyle{acl_natbib}

\end{document}

%% file: 1_introduction.tex
\section{Introduction}
\label{sec:intro}
\begin{figure}[ht!]
    \includegraphics[width=\columnwidth, height=5cm,keepaspectratio]{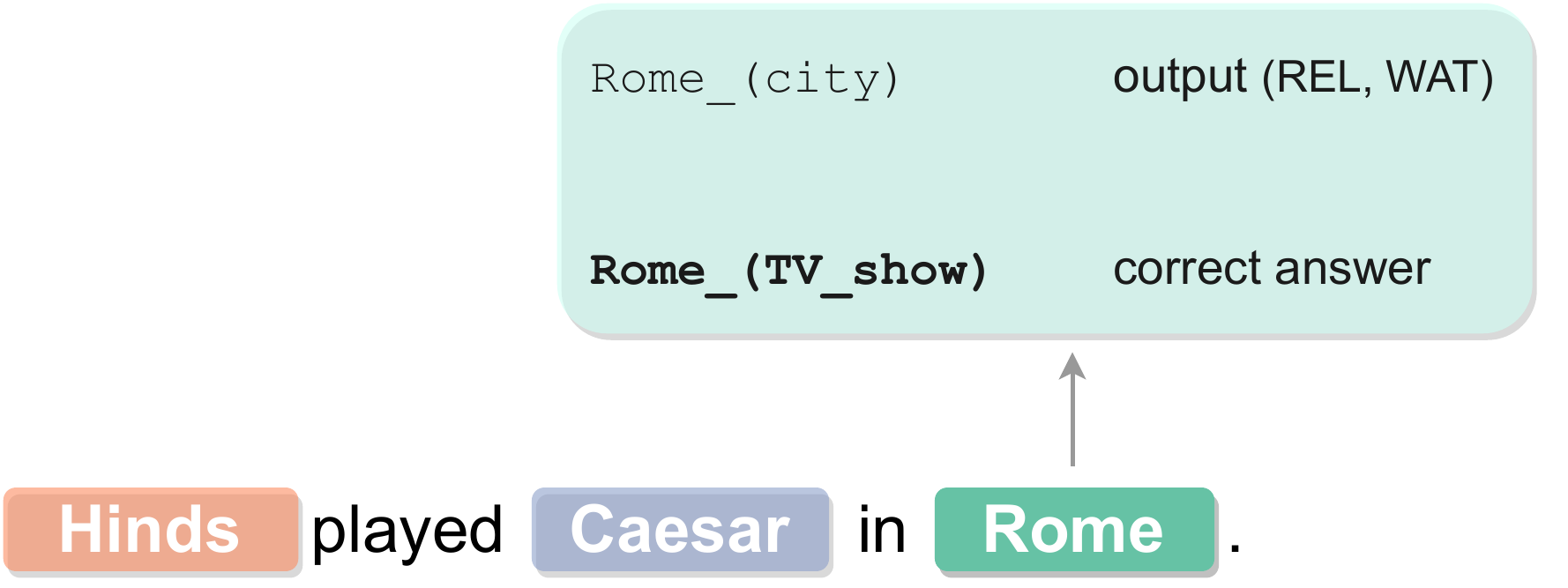}
    \caption{An example of entity overshadowing: two popular ED systems (REL and WAT) predict the most frequent entity (Rome the city) instead of the correct answer in this context (Rome the TV series).} 
    \label{fig:overshadowing_example}
\end{figure}

Entity disambiguation (ED) is the task of mapping an ambiguous entity mention to the corresponding entry in a structured knowledge base. Despite ED being a well-known task, recent work has shown that the existing methods are still far from achieving human-level performance: in particular, the case of entity overshadowing remains a big challenge. An entity $e_1$ overshadows $e_2$ if the two entities share the same surface form $m$, and $e_1$ is more common than $e_2$, i.e., has a higher prior probability to be linked to $m$ ~\citep{provatorova2021robustness}.  For example, when given the sentence \textit{“\underline{Michael Jordan} published a paper on machine learning”} and the task of linking \textit{Michael Jordan} either to the basketball player (a frequent entity) or to the scientist (an overshadowed entity), a human will correctly choose the latter, while a typical model is likely to ignore the context and give the wrong yet more popular answer due to over-relying on prior probability. Figure~\ref{fig:overshadowing_example} shows another example of entity overshadowing: the entity Rome (TV series) is overshadowed by Rome (city). 

According to previous research, current ED systems are prone to over-relying on prior probability instead of focusing on context information, which causes them to underperform on overshadowed entities. In benchmarking experiments performed by~\citet{provatorova2021robustness}, all ED systems under evaluation appeared to have a large performance gap between Top and Shadow subsets of the ShadowLink dataset, where Top contains most frequent entities and Shadow contains their overshadowed counterparts. The results of a human evaluation experiment in the same study indicate that the challenge of entity overshadowing is unique to automated ED methods: human participants achieved equally good results at disambiguating entities sampled from Top and Shadow. These findings call for further research in the field of ED, with the goal of building a method that outperforms existing systems on overshadowed entities while still achieving competitive results on standard datasets.


Interestingly, the best results on Shadow in the benchmarking experiments were achieved by AIDA~\citep{hoffart2011robust}, an unsupervised collective entity disambiguation method: while still affected by overshadowing, this method appeared to be the best at capturing the context information in comparison with modern neural appproaches. Specifically, AIDA relies on two main sources of context information: semantic similarity between an entity and its context and graph-based relatedness between the candidate entities of different mentions. Our study continues this line of work, incorporating modern neural methods to measure semantic similarity and adding novel heuristics to improve candidate filtering and collective disambiguation.
%

We introduce NICE (NER\footnote{Named Entity Recognition~\cite{yadav-bethard-2018-survey}}-enhanced Iterative Combination of Entities), a combined entity disambiguation algorithm designed to tackle the challenge of entity overshadowing by focusing on three aspects of context-based information: entity types, entity-context similarity and entity coherence. 
The pipeline of NICE includes a NER-enhanced candidate filtering module designed to improve robustness on overshadowed entities (Section \ref{sec:candidate_filtering}), a pre-scoring module that calculates semantic similarity between a candidate entity and a mention in context, and an unsupervised iterative disambiguation algorithm that maximises entity coherence (Section \ref{sec:collective_disambiguation}), combining the relatedness scores between candidate entities with the scores of the semantic similarity module (Sections \ref{sec:collective_disambiguation}-\ref{sec:params}).
To the best of our knowledge, our study is the first attempt to build an entity disambiguation method designed specifically to tackle the problem of entity overshadowing.

We perform a systematic evaluation of the NICE method, and use our experimental results to answer the following research questions:

\textbf{RQ1:} Does focusing on context information improve ED performance on overshadowed entities?

\textbf{RQ2:} Does focusing on context information instead of relying on mention-entity priors in ED allow to maintain competitive performance on more frequent entities?
 
\textbf{RQ3:} In what ways do the different aspects of context information contribute to ED performance on overshadowed entities?

We hope that our work will encourage further studies concerning overshadowed entities. The source code of the NICE method is provided as supplementary material and will be released publicly upon acceptance.

%% file: 2_methods.tex
\section{The NICE method}
\begin{figure*}[ht!]
    \centering
    \includegraphics[width=2\columnwidth, height=4.5cm,keepaspectratio]{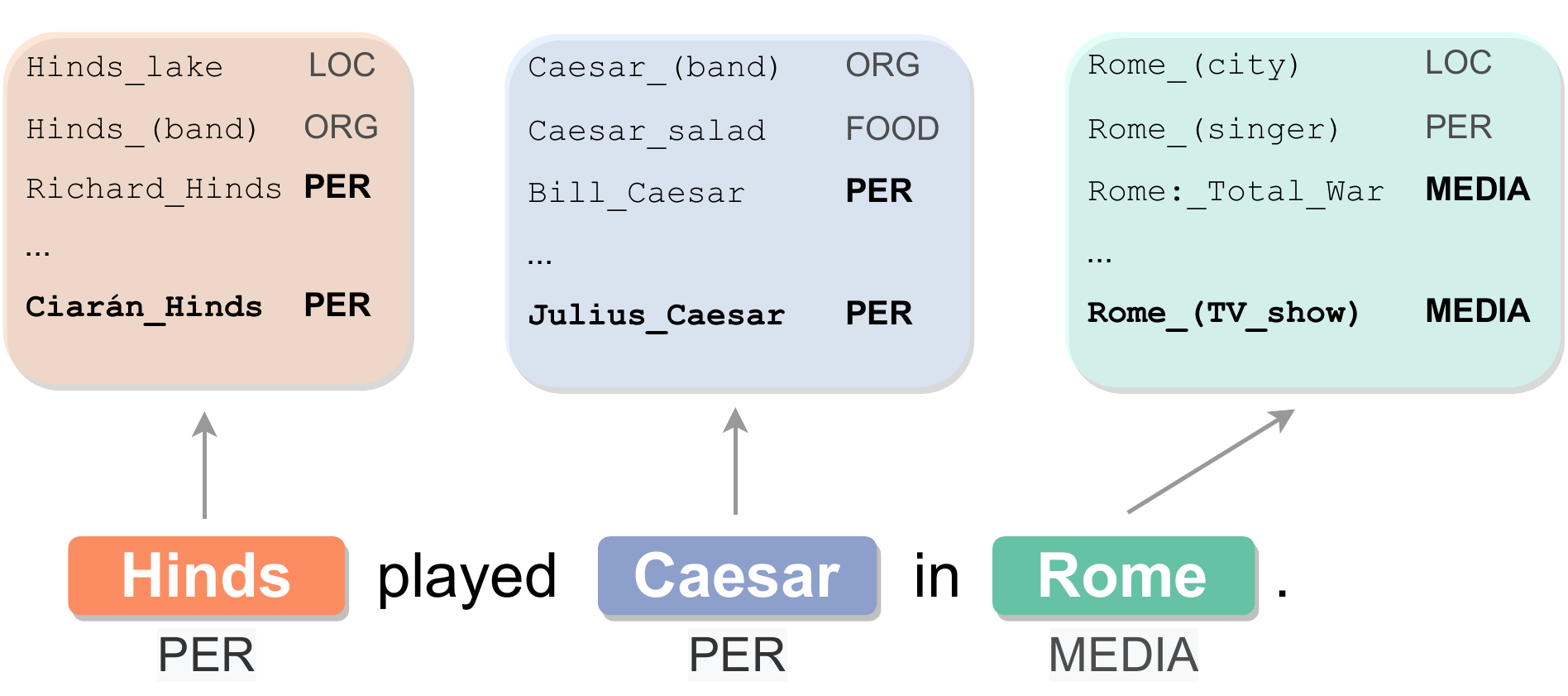}
    \caption{An example of filtering candidate entities by their NER types. A candidate entity is selected (highlighted in bold) if its NER type coincides with the predicted NER type of the corresponding entity mention. For simplicity, we assume that the confidence score is always higher than the threshold, so that only the first predicted type is considered for filtering.} 
    \label{fig:ner_types_example}
\end{figure*}
Our method is based on the assumption that the main challenge in disambiguating overshadowed entities stems from over-relying on entity commonness, and therefore switching the focus to the context (entity relatedness) can improve the performance. We consider three main ways of extracting information from the context: (1) using mention-entity similarity to predict entity types and improve candidate filtering, (2) using word embeddings enhanced with entity types to measure semantic similarity between an entity and its context, and (3) using entity-entity similarity to make sure that the entity disambiguating decisions within one document are coherent (collective disambiguation). 
\subsection{Candidate filtering} \label{sec:candidate_filtering} Adding the step of filtering candidate entities before disambiguation brings the benefits of reduced inference time and potential improvements in accuracy. To perform this step in the NICE method, we follow the work of~\citet{tedeschi2021named} by using entity type information. Given an entity mention $m$ surrounded by textual context $(cont_{left}, cont_{right})$ and a list of candidate entities ${cands} = \{e_1, \dots., e_n\}$, we use a NER classifier to predict the top-k possible entity types of $m$. Then, we discard all candidate entities that have an entity type not matching any of these $k$ classes: $$cands_{filtered} = \{e_i: type(e_i) \in \hat{T} \;|\; e_i \in {cands}\},$$ where $\hat{T}$ is the set of top-k predicted entity types. If the confidence score of the NER classifier is above a threshold value $t$, only one class is used instead of $k$. In the current setup of the NICE method, the number of top predicted classes is $k=3$ and the confidence threshold value is $t=1$, which means that the classifier always outputs the top-3 entity classes. 
Figure~\ref{fig:ner_types_example} shows an example of NER-based candidate filtering.

To obtain the entity types for the candidates, we use the Wiki2NER dictionary provided by~\citet{tedeschi2021named}\footnote{\url{https://github.com/Babelscape/ner4el}}. Then, instead of using the NER classifier as provided by \citet{tedeschi2021named}, which has been trained only on the AIDA training set and therefore may be biased towards frequent entities as well, we introduce a refined version of it, which is more robust to overshadowing. Specifically, we filter the training set of BLINK~\citep{wu2020scalable}\footnote{BLINK is a dataset for ED consisting of 9M entries extracted from Wikipedia.} by discarding the entries where the ground truth answer has the highest popularity score among all candidate entities
. Then, we use the 2M remaining data entries to fine-tune the classifier. The motivation behind fine-tuning the classifier rather than training it from scratch is to achieve an improvement in recognising overshadowed entities without losing the knowledge about more popular entities. This way, we obtain a system that specialises on overshadowed entities while still performing well in a standard ED setup. 
\subsection{Candidate pre-scoring} To obtain relevance scores for the candidate entities before the collective disambiguation phase, NICE relies on semantic similarity between a candidate entity and its corresponding mention in context.
To calculate these semantic similarity scores, we use the NER-enhanced word embeddings produced by the NER4EL model~\citep{tedeschi2021named}. This model uses a dual-encoder Transformer-based architecture that, given as input a mention-candidate pair $\langle m, c \rangle$, produces a vector representation for both $m$ and $c$. Then, the relatedness between the two vectors is measured as the cosine similarity score. To encode the mention, the model uses both the mention itself and its context as input. To encode the candidate entities, instead, the model uses their textual descriptions from Wikipedia.
\subsection{Collective disambiguation}\label{sec:collective_disambiguation} We follow the assumption that the entities within one document are coherent, i.e. there exists a coherence measure $C$ such that $C(e^{*}_1, ..., e^{*}_n) \geq C(e_1, ..., e_n)$ where $e^{*}_1, ..., e^{*}_n$ are the correct entities for mentions $m_1, ..., m_n$, and $e_1, ..., e_n$ is any other selection of candidate entities for these mentions (with one candidate per mention). 
Then, we use the following heuristic to reduce the search space and speed up the disambiguation process: if the entities within one document are coherent, and different mentions have different degrees of ambiguity (different sizes of candidate sets), then using an iterative process that starts with the least ambiguous mention may reduce the chance of wrong disambiguation decisions \cite{barba2021consec}. 

We thus propose ICE (for "Iteratively Combining Entities"), a collective disambiguation algorithm  based on the iterative disambiguation process described in Algorithm~\ref{alg:ice}. The input of the algorithm is a set of entity mentions within one document, where every mention is matched with a pre-scored list of candidate entities. The first step of the disambiguation process consists of selecting a seed mention: the least ambigous entity mention, i.e., one that has the lowest number of candidates. The seed mention is disambiguated without using entity relatedness information, relying on the input scores instead. Then, an iterative process begins: on every step of the algorithm the least ambiguous entity mention is selected and removed from the set of unprocessed mentions. This mention is disambiguated based on coherence: for every candidate entity, the algorithm calculates an aggregated relatedness score between the entity and the set of already disambiguated entities, and chooses the candidate that maximises this score. The algorithm is terminated when the set of unprocessed entity mentions is empty.
\begin{algorithm}[!t]
    \caption{ICE disambiguation}
    \label{alg:ice}
\small
\setstretch{1.2}
    \SetKwInOut{Input}{Input}
    \SetKwInOut{Output}{Output}
    \SetKwInOut{Where}{~}
	\DontPrintSemicolon
    \Input{$S_{todo} = \{{ment}_i: cands_i, scores_i\}_{i=1}^n$ \; where~$cands_i = [{cand}^1_i, ..., {cand}^{k_i}_i]$ \; and $scores_i = [{score}^1_i, ..., {score}^{k_i}_i]$} 
    \Output{$S_{ans} = \{{ment}_i:  {cand}^*_i\}_{i=1}^n$}
    \tcp{Part 1. Seed entity disambiguation}
        $m_{seed} \leftarrow find\_least\_ambiguous(S_{todo})$\;
        $S_{todo} \leftarrow S_{todo} \setminus \{m_{seed}, cands_{seed}\}$ \;
        $cand_{seed} \leftarrow disamb\_seed(m_{seed}, cands_{seed})$ \;
        $S_{ans} \leftarrow S_{ans} \cup \{m_{seed}, cand^*_{seed}\}$ \;
    \tcp{Part 2. Collective disambiguation}
        \While{$S_{todo} \neq \emptyset$} {
          $m_i \leftarrow find\_least\_ambiguous(S_{todo})$ \;
         $S_{todo} \leftarrow S_{todo} \setminus \{m_i, cands_i\}$ \;
         \tcp{Choose the candidate closest to the already disambiguated entities}
         $cand^*_i \leftarrow argmax~aggr\_rel(cands_i, scores_i, S_{ans})$ \;
         $S_{ans} \leftarrow S_{ans} \cup \{m_i, cand^*_i\}$
    }
\end{algorithm}
Note that the relatedness score in Algorithm~\ref{alg:ice} can be calculated and aggregated in different ways. In the setup used in the NICE method, we calculate the score of each candidate as a weighted average between the entity coherence score and the original input score:\\

\noindent$score_{final} = \alpha score_{coherence} + (1-\alpha) score_{input}$.\\

The process of choosing the parameter values is described in detail in the next subsection.
\subsection{Parameters of the method}\label{sec:params}
The NICE method has four main parameters to experiment with: confidence threshold $t$ of the candidate filtering model, weight $\alpha$ of the ICE coherence score, the relatedness measure used in collective disambiguation and the relatedness aggregation method.
To find the best combination of the parameters, we sampled a development set from the training data of ShadowLink and used grid search. The development set contains 100 Top and 100 Shadow entities: the two classes are represented equally because a system cannot distinguish between Top and Shadow on the inference step. 
For the confidence threshold value $t$, we considered the values from 0.5 to 0.9 with the step 0.1. For the ICE score weight $\alpha$, the search space consisted of the values between 0 and 1 with the step 0.1. 

To find the most suitable entity relatedness measure, we experimented with the seven measures available in the WAT relatedness API~\footnote{\url{https://sobigdata.d4science.org/web/tagme/wat-api}}~\citep{piccinno2014tagme}, which retrieves relatedness scores by Wikipedia IDs of the corresponding entity pages. The measures available in the API are the following: PMI (Pointwise Mutual Information), Milne-Witten, LM (language model), WordVec, Jaccard, Barabasi-Albert, and conditional probability. 

For aggregation of the relatedness scores in Algorithm~\ref{alg:ice}, we considered three setups: \textit{max}, \textit{min} and \textit{avg}. The \textit{max} setup is a greedy approach, where the score of a candidate entity is calculated as its maximum pairwise relatedness to the already disambiguated entities. It relies on the assumption that a relevant entity does not necessarily need to be close to \textbf{all} other entities in the document: it is enough to have a high pairwise relatedness with just one other entity. The opposite setup is \textit{min}: it follows a strict assumption that all entities within one document must be coherent, and calculates the aggregated score as the minimum pairwise relatedness to the already disambiguated entities. Finally, the \textit{avg} setup calculates the score as a mean pairwise relatedness between a candidate and the disambiguated entities.
\begin{figure}[t!]
    \centering
    \includegraphics[width=\columnwidth, height=6.3cm,keepaspectratio]{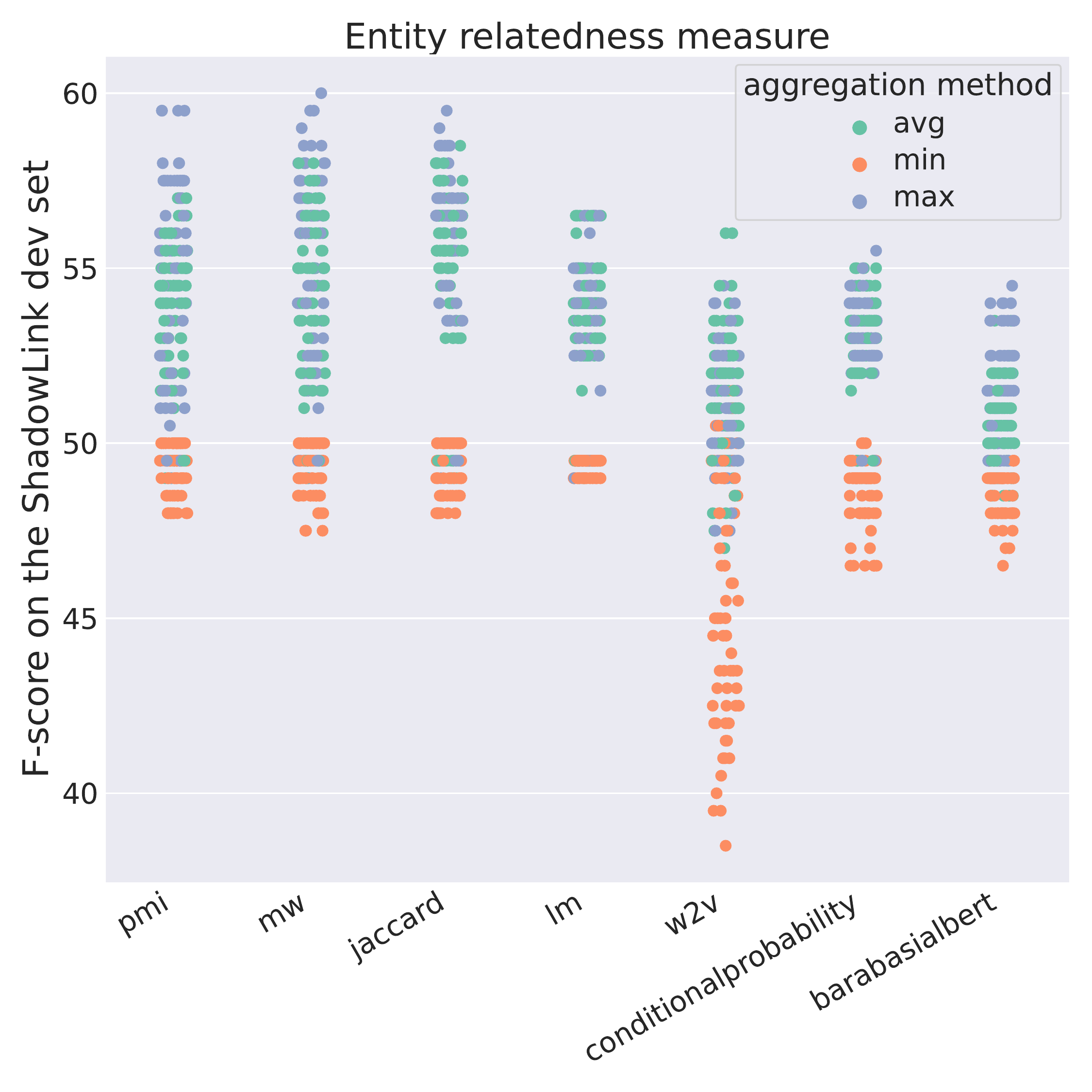}
    \caption{Micro F-score values achieved on the ShadowLink dev set with different combinations of relatedness measures and aggregation methods.}
    \label{fig:relatedness_measures}
\end{figure}
Figure~\ref{fig:relatedness_measures} shows the scores achieved on the development set by using the combinations of the seven relatedness measures and three aggregation setups. One point on the figure represents the score achieved with one combination of the four parameters of NICE. The best relatedness measure turned out to be Milne-Witten~\citet{10.1145/1458082.1458150}:
$$MW(e_1, e_2) =\frac{\log\frac{\max(|L_{e_1}|, |L_{e_2}|)}{(|L_{e_1}\cap L_{e_2}|)}}{\log\frac{|W|}{\min(|L_{e_1}|, |L_{e_2}|)}},$$
 where $L_{e_i}$ is the set of Wikipedia pages that link to the page of entity $e_i$, and $W$ is the entire Wikipedia. 
 The best aggregation method according to our experiments is taking the maximum value of relatedness:
 
 \begin{center}
     $s(e_i) = max\{{MW}(e_i, \hat{e}_j) | \hat{e}_j \in S_{ans}\}$,
 \end{center}
 
 \noindent where $ S_{ans}$ is the set of entities already processed by the algorithm and $e_i$ is a candidate entity.
 
\begin{figure}[t!]
    \includegraphics[width=\columnwidth, height=5.7cm,keepaspectratio]{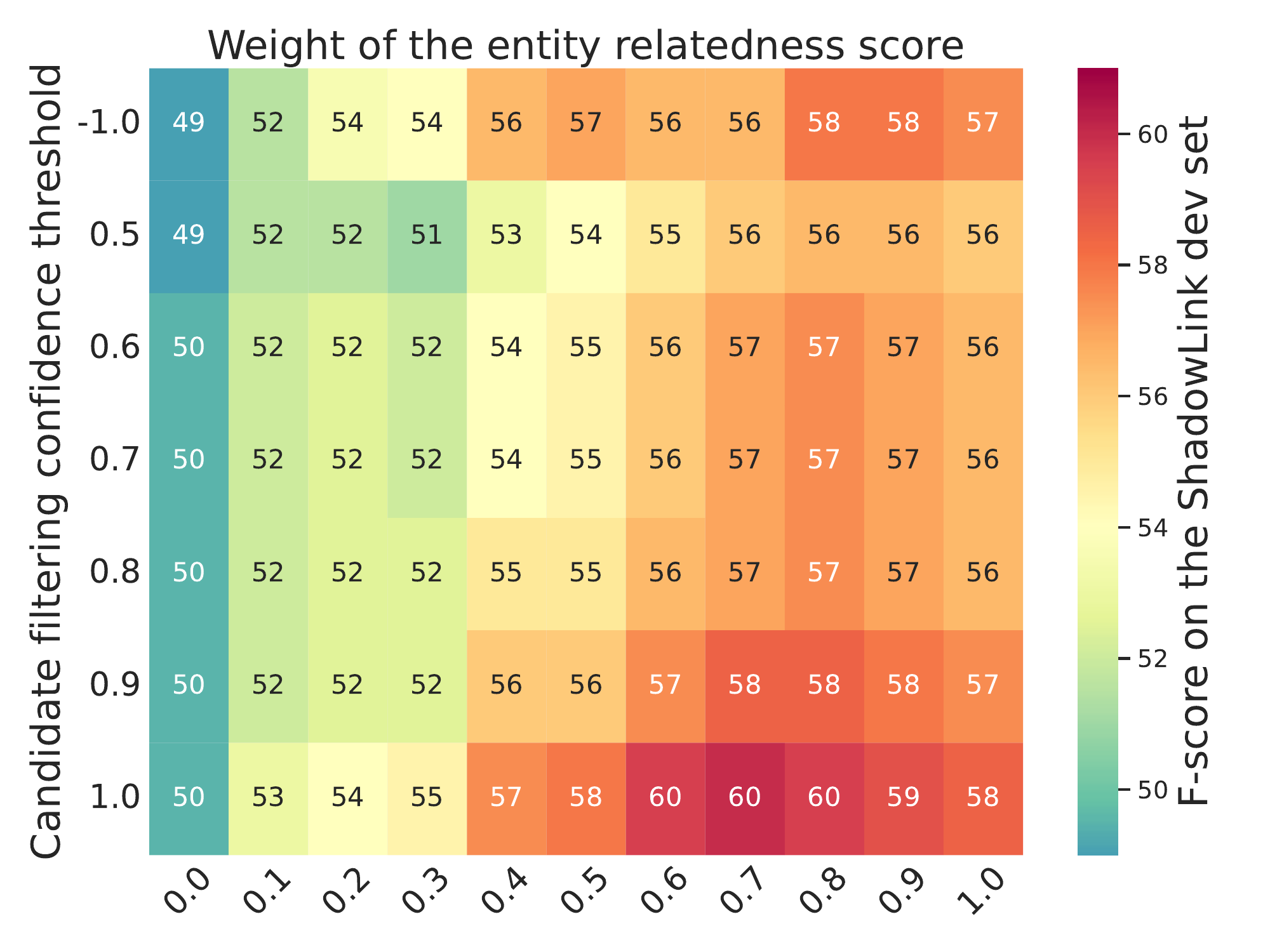}
    \caption{Micro F-score values achieved on the ShadowLink dev set using Milne-Witten entity similarity measure and \textit{max} aggregation method.}
    \label{fig:heatmap_mw}
\end{figure}
Figure~\ref{fig:heatmap_mw} demonstrates a heatmap of F-scores achieved on the ShadowLink development set with different combinations of candidate filtering threshold values and relatedness score weights while using the Milne-Witten relatedness measure and the \textit{max} aggregation method. The value -1 of the filtering confidence threshold indicates not using the candidate filter at all. From the figure it can be seen that the best combination of the two remaining parameters is the confidence value of 1 (which means always predicting top-3 entity classes) and the weight 0.7 assigned to the relatedness score, which corresponds to the weight 0.3 of the semantic similarity score. Thus, the final score of each candidate entity is calculated as:
\begin{flalign*}
\centering
s(e_i) & = 0.7 \max\{{MW}(e_i, \hat{e}_j) | \hat{e}_j \in S_{ans}\} & \\&\;\;\;\; + 0.3 s_{input}(e_i).
\end{flalign*}
\subsection{Implementation details}
Our contribution is focused on the task of entity disambiguation, where the mention boundaries and the candidate entities are given. We note that different methods of mention detection and candidate generation can be combined with the NICE disambiguation module to perform end-to-end entity linking. To perform collective disambiguation on ShadowLink, where only one entity span per entry is available, we used TagMe API~\footnote{\url{https://sobigdata.d4science.org/web/tagme/tagme-help}}~\citep{DBLP:conf/cikm/FerraginaS10} for mention detection and REL\footnote{\url{https://github.com/informagi/REL}}~\citep{van2020rel} for candidate generation.

%% file: 3_experiments.tex
\section{Experimental Setup}
In this section, we describe the datasets we use to evaluate our methods (Section \ref{sec:datasets}) and the baseline systems we compare with (Section \ref{sec:baselines}).

\subsection{Datasets}\label{sec:datasets}
As the NICE method is designed specifically for improving disambiguation of overshadowed entities, the main dataset for evaluating its performance is
ShadowLink~\citep{provatorova2021robustness}. ShadowLink contains ambiguous entities with short textual context gathered from the Web: it includes the Top subset which contains popular entities, and the Shadow subset which contains overshadowed entities, where every Shadow entry is matched with one Top entry by a shared entity surface form. 
To make sure that the main challenge is entity disambiguation and not candidate generation, we excluded all entries for which the candidate generation module of REL failed to retrieve the correct entity. The remaining dataset contains 491 Shadow and 614 Top entries.

Note that, by design, ShadowLink contains ground truth information for only one entity mention per entry. As our method relies on collective disambiguation, we extract all different entity mentions from every entry and use them in the inference – however, the evaluation is only performed on the "target mention" that has a ground truth label assigned to it. 
To test the collective disambiguation approach for multiple mentions within one document, as well as to make sure that our method performs competitively on standard data, we also evaluate NICE on the test subset of the AIDA-YAGO-CoNLL dataset~\citep{hoffart2011robust}, the largest manually annotated ED evaluation set which consists of 388 news articles.
\subsection{Baselines}\label{sec:baselines} We compare the NICE method with five popular ED models available in the GERBIL evaluation framework \citep{roder2018gerbil}: AIDA~\citep{hoffart2011robust}, DBpedia Spotlight~\citep{DBLP:conf/i-semantics/MendesJGB11}, AGDISTIS/MAG~\citep{usbeck2014agdistis}, Babelfy~\citep{moro2014entity} and WAT~\citep{piccinno2014tagme}, as well as three novel models not yet available in GERBIL: REL~\citep{van2020rel}, GENRE~\citep{de2020autoregressive}, and NER4EL~\citep{tedeschi2021named}. 
%

The newest of the systems under evaluation is NER4EL, a neural model that uses the information about entity types to make ED decisions based on semantic similarity between a candidate entity and its context.
As NER4EL turns out to achieve high results on Shadow, we include it as the semantic similarity module in the combined NICE method. Combining NER4EL with the ICE algorithm allows to extract more information from the context than by using each of the two methods alone, as NER4EL does not include entity relatedness information and ICE does not include semantic similarity.

%% file: 4_big_table.tex
\begin{table*}[ht!]
\addtolength{\tabcolsep}{10pt}
\centering
\begin{tabular}{@{}lrrr@{}}
\toprule
\textbf{Method} & \textbf{Shadow filtered} & \textbf{Top filtered} & \textbf{AIDA test} \\
\midrule
AIDA~\citep{hoffart2011robust} & 45.4 & 65.0 & 81.8\\ 
DBpedia Spotlight~\citep{DBLP:conf/i-semantics/MendesJGB11} & 16.2 & 37.9 & 49.3\\
Babelfy~\citep{moro2014entity} & 52.9 & 66.1 & 68.2\\
WAT~\citep{piccinno2014tagme} & 32.9 & 59.5 & 60.7\\
AGDISTIS/MAG~\citep{usbeck2014agdistis} & 12.8 & 22.8 & 59.4\\
REL~\citep{van2020rel} & 31.5 & 70.5 & 83.3\\
GENRE~\citep{de2020autoregressive} & 33.8 & 54.2 & \textbf{93.3}\\
NER4EL~\citep{tedeschi2021named} & 44.3 & 62.4 & \underline{92.5}\\
\midrule
NER4EL + no candidate filtering & 43.3 & 67.3 & 86.3\\%
NER4EL + robust candidate filter & 45.7 & 68.1 & 85.3\\
ICE + no candidate filtering & 56.9 & 74.9 & 72.1\\%
ICE + robust candidate filter & \textbf{59.9} & \underline{76.1} & 71.9\\
\midrule
NICE & \underline{58.3} & \textbf{77.2} & 80.4 \\
\bottomrule
\end{tabular}%
\caption{\label{tbl:res-all} Benchmark evaluation results in terms of micro-$F_1$. In the top part of the table, we report the results of the baseline systems we compare with. In the bottom and middle parts, we report the scores obtained by the NICE method and its intermediate versions, respectively. We mark in \textbf{bold} the best scores and \underline{underline} the second best.}
\end{table*}

%% file: 4_results.tex
\section{Results and discussion}
The results of our experiments are presented in Table~\ref{tbl:res-all}. Note that the term "filtered" in the dataset description refers to filtering out the entries where the correct candidate entity was not retrieved by our candidate generation model, and not to the NER-based candidate filtering. 
We use the experimental data to answer the research questions introduced in Section~\ref{sec:intro}.

\paragraph{RQ1:} \textit{Does focusing on context information improve ED performance on overshadowed entities?}

From the evaluation results in Table~\ref{tbl:res-all} it can be seen that NICE outperforms all baseline systems on the Shadow subset by a large margin. Moreover, all variations of our method considered in the experiments achieve top results on Shadow, which shows that all ways of leveraging context information considered in our study lead to performance improvements on overshadowed entities.

\paragraph{RQ2:} \textit{Does focusing on context information instead of relying on mention-entity priors in ED allow to maintain competitive performance on more frequent entities?}

Our experimental data shows that NICE outperforms all baselines on the Top subset of ShadowLink, where every entry contains the most frequent entity from the corresponding entity space. This demonstrates that placing more impact on context information does not automatically decrease the performance on more frequent entities: on the contrary, it appears to be beneficial in the setup of using the short textual context of ShadowLink.

The results on the AIDA dataset, however, are considerably lower compared to several of the baselines. There are different factors contributing to this performance drop. Firstly, it appears that the original NER classifier used for candidate filtering in NER4EL works much better on the AIDA dataset than our classifier, which has been trained to focus on overshadowed entities. Secondly, the input structure differed between the two dataset types: for Top and Shadow we extracted auxiliary mention spans from the context to perform collective disambiguation (as the ShadowLink dataset only contains one ground truth entity mention per entry), and for AIDA we only used the mention spans provided in the dataset. In the case of AIDA, all the labelled mentions satisfy the strict definition of a named entity, while for ShadowLink the auxiliary mentions were noun phrases extracted using the TagMe API, which considers all Wikipedia anchors as entities. For example, in the sentence \textit{"Michael Jordan published a paper"}, \textit{paper} is not a named entity but a common concept that can be linked to Wikipedia and used for collective disambiguation. 
Interestingly, a similar idea is used by Babelfy~\citep{moro2014entity}, which disambiguates named entities and common concepts simultaneously using graph information, achieving the second-highest score both on Top and Shadow datasets. This indicates that considering all linkable mentions within a short textual context is benificial for disambiguating overshadowed entities.

\paragraph{RQ3:} \textit{In what ways do the different aspects of context information contribute to ED performance on overshadowed entities?}

NICE uses three aspects of context information: entity types, semantic similarity between a candidate entity and its mention in context, and graph-based relatedness between candidate entities. To estimate the impact of these aspects, we turn to the experimental data of two kinds: the evaluation results presented in Table~\ref{tbl:res-all} and the results achieved on the development set with different combinations of component weights, shown in Figure~\ref{fig:heatmap_mw}.

From Table~\ref{tbl:res-all} it can be seen that using entity types is beneficial both for candidate filtering and for enhancing the semantic similarity module: NER4EL achieves considerably high results on Shadow when used alone and demonstrates a further improvement when combined with the robust candidate filter.  Figure~\ref{fig:heatmap_mw} shows that the highest scores are achieved when the weight of graph-based entity coherence score is relatively high, with the best combination being $0.3 score_{NER4EL} + 0.7 score_{ICE}$. 

Moreover, the best results on the Shadow test set are achieved when using the ICE coherence score alone (middle part of Table~\ref{tbl:res-all}). Thus, all aspects of context information are beneficial for disambiguating overshadowed entities, with the entity coherence being the most important component.

%% file: 5_related_work.tex
\section{Related work}
The task of entity linking in its modern sense has been first introduced in 2007 by~\citet{10.1145/1458082.1458150}. Since then, multiple evaluation datasets have been proposed, and numerous methods have been developed to improve the results. 
The first methods of entity disambiguation focused explicitly on two components: entity \textit{commonness}, which shows how likely a particular entity is to be linked to a given mention regardless of the context, and entity \textit{relatedness}, which shows how relevant an entity is to a given context~\citep{10.1145/1321440.1321475}. Three most prominent methods that used this approach are AIDA~\citep{hoffart2011robust}, TagMe~\citep{DBLP:conf/cikm/FerraginaS10} and WAT~\citep{piccinno2014tagme}. AIDA~\citep{hoffart2011robust} uses a graph-based collective disambiguation algorithm enhanced with robustness tests, building weighted edges between mentions and candidate entities as well between different candidate entities, and removing an edge on every iteration to maximise the minimum weighted degree of the graph. TagMe~\citep{DBLP:conf/cikm/FerraginaS10}, instead, uses the relatedness measure defined by \citet{10.1145/1458082.1458150} weighted with the commonness of a sense together with the keyphraseness measure defined by \citet{10.1145/1321440.1321475} to exploit the context around the target word. Finally, WAT \citep{piccinno2014tagme} is a redesigned
system of TagMe and includes graph-based algorithm for ranking entities in entity graph based on entity relatedness, and vote-based algorithm for local disambiguation. 
Another approach to entity disambiguation presented in the same year as WAT is Babelfy~\citep{moro2014entity}, a method that combines entity linking with word sense disambiguation, using an algorithm based on random walks on the BabelNet multilingual graph to disambiguate all linkable mentions together with senses of words occurring in the text.

With the advantage of deep learning methods in NLP, new approaches have been proposed to improve the performance of ED. \citet{van2020rel} introduced REL, a modular system that combines several of these approaches: candidate generation with mention-entity priors proposed by~\citet{ganea2017deep}, entity disambiguation with mention-wise normalisation first introduced by~\citet{le2018improving}, and entity-context similarity computed with Wikipedia2Vec embeddings by~\citet{yamada2016joint}. While achieving top performance on standard benchmarks, REL appears to suffer from over-relying on its prior predictions, which leads to a considerable performance drop on overshadowed entities.
\citet{de2020autoregressive}, instead, proposed a novel autoregressive approach to entity linking that, given a mention in context, generates the title of the corresponding Wikipedia
page.

While learning from large amounts of data allows the modern neural ED methods to achieve high performance on most of the evaluation datasets, it also puts them at risk of overfitting to the most frequently seen entities and overlooking the important edge cases.  \citet{ilievski2018systematic} showed that entities with low frequency and high ambiguity appear to be underrepresented in standard ED evaluation datasets, despite such entities being especially challenging to disambiguate.~\citet{provatorova2021robustness} continued this line of work, introducing the concept of entity overshadowing and releasing ShadowLink, the first dataset designed specifically to study this phenomenon. Both studies conclude that the problem of entity disambiguation is still far from solved, and call for ED systems to consider more complex cases instead of only optimising for the standard evaluation frameworks. \citet{provatorova2021robustness} mentioned specifically that relying more on context information and less on priors appears to be the key to achieving better performance on overshadowed entities. 

\citet{tedeschi2021named} proposed NER4EL, a neural ED method that leverages the context by using information about entity types on four levels: candidate filtering, entity representation, negative sampling during training, and entity decoding. This allows NER4EL to perform on par with~\citet{de2020autoregressive} despite only training on a small fraction of data used by GENRE. While not aiming specifically to improve performance on overshadowed entities, NER4EL achieves high results on the ShadowLink dataset in our study. One limitation of the method is its lack of a collective entity disambiguation method: every mention within a document is disambiguated separately, using only its textual context. By combining NER4EL with the ICE collective disambiguation algorithm introduced in this paper, we manage to achieve top results on overshadowed entities.

%% file: 6_conclusion.tex
\section{Conclusion}
We introduced NICE, an overshadowing-aware entity disambiguation method that consists of three main components: a candidate filtering module designed for improved performance on overshadowed entities, a collective disambiguation module that uses an unsupervised iterative algorithm to capture entity coherence, and a module that measures semantic similarity between an entity and its context using NER-enhanced word embeddings. Our experimental results show that NICE achieves substantial improvements on overshadowed entities compared to the baseline methods, while still performing competitively in a standard entity disambiguation setting. This demonstrates that encompassing explicit contextual features, such as entity types and entity coherence, improves ED performance on overshadowed entities.

Future work directions include using different annotation methods for extracting linkable concepts (not necessarily named entities) from the textual context, as well as experimenting with longer textual contexts on the extended version of ShadowLink and other ED datasets.

%% file: 7_limitations.tex
\section{Limitations}
While the NICE method demonstrates top performance on overshadowed entities, its current implementation includes several limitations. Firstly, the method relies on an external web API to calculate entity relatedness, which requires the user to register and obtain an access token.\footnote{\url{https://sobigdata.d4science.org/web/tagme/tagme-help}} Secondly, due to computational limitations our method only considers 15 candidates for calculating semantic similarity scores. Adding more candidates may lead to performance improvements. 
Thirdly, while the ICE algorithm attempts to minimise the runtime by processing the least ambiguous entities first, there could be a possible edge case where all the input mentions have the same high number of candidates. In this case, the time complexity will be $O(k N^2)$, where $k$ is the number of candidates and $N$ is the number of mentions, which slows down the disambiguation process, especially when using a web API for calculating relatedness. 
Lastly, while NICE outperforms half of the baselines on the AIDA dataset, it achieves considerably lower scores than the two newest models, GENRE and NER4EL. We assume that the main reason behind it is not using all linkable noun phrases from the textual context and only relying on the mention spans provided in the input, contrary to the setup used on ShadowLink. Further research is needed to test this assumption and experiment with extracting additional mentions from the AIDA dataset. 

%% file: main.bbl
\begin{thebibliography}{20}
\expandafter\ifx\csname natexlab\endcsname\relax\def\natexlab#1{#1}\fi

\bibitem[{Barba et~al.(2021)Barba, Procopio, and Navigli}]{barba2021consec}
Edoardo Barba, Luigi Procopio, and Roberto Navigli. 2021.
\newblock \href {https://aclanthology.org/2021.emnlp-main.112/} {Consec: Word
  sense disambiguation as continuous sense comprehension}.
\newblock In \emph{Proceedings of the 2021 Conference on Empirical Methods in
  Natural Language Processing}, pages 1492--1503.

\bibitem[{De~Cao et~al.(2020)De~Cao, Izacard, Riedel, and
  Petroni}]{de2020autoregressive}
Nicola De~Cao, Gautier Izacard, Sebastian Riedel, and Fabio Petroni. 2020.
\newblock \href {https://arxiv.org/abs/2010.00904} {Autoregressive entity
  retrieval}.
\newblock In \emph{International Conference on Learning Representations}.

\bibitem[{Ferragina and Scaiella(2010)}]{DBLP:conf/cikm/FerraginaS10}
Paolo Ferragina and Ugo Scaiella. 2010.
\newblock \href {https://dl.acm.org/doi/abs/10.1145/1871437.1871689} {Tagme:
  on-the-fly annotation of short text fragments (by wikipedia entities)}.
\newblock In \emph{Proceedings of the 19th ACM international conference on
  Information and knowledge management}, pages 1625--1628.

\bibitem[{Ganea and Hofmann(2017)}]{ganea2017deep}
Octavian-Eugen Ganea and Thomas Hofmann. 2017.
\newblock \href {https://doi.org/10.18653/v1/D17-1277} {Deep joint entity
  disambiguation with local neural attention}.
\newblock In \emph{Proceedings of the 2017 Conference on Empirical Methods in
  Natural Language Processing}, pages 2619--2629, Copenhagen, Denmark.
  Association for Computational Linguistics.

\bibitem[{Hoffart et~al.(2011)Hoffart, Yosef, Bordino, F{\"u}rstenau, Pinkal,
  Spaniol, Taneva, Thater, and Weikum}]{hoffart2011robust}
Johannes Hoffart, Mohamed~Amir Yosef, Ilaria Bordino, Hagen F{\"u}rstenau,
  Manfred Pinkal, Marc Spaniol, Bilyana Taneva, Stefan Thater, and Gerhard
  Weikum. 2011.
\newblock \href {https://aclanthology.org/D11-1072/} {Robust disambiguation of
  named entities in text}.
\newblock In \emph{Proceedings of the 2011 Conference on Empirical Methods in
  Natural Language Processing}, pages 782--792.

\bibitem[{van Hulst et~al.(2020)van Hulst, Hasibi, Dercksen, Balog, and
  de~Vries}]{van2020rel}
Johannes~M. van Hulst, Faegheh Hasibi, Koen Dercksen, Krisztian Balog, and
  Arjen~P. de~Vries. 2020.
\newblock \href {https://arxiv.org/abs/2006.01969} {{REL:} an entity linker
  standing on the shoulders of giants}.
\newblock In \emph{Proceedings of the 43rd International {ACM} {SIGIR}
  conference on research and development in Information Retrieval, {SIGIR}
  2020, Virtual Event, China, July 25-30, 2020}, pages 2197--2200.

\bibitem[{Ilievski et~al.(2018)Ilievski, Vossen, and
  Schlobach}]{ilievski2018systematic}
Filip Ilievski, Piek Vossen, and Stefan Schlobach. 2018.
\newblock \href {https://aclanthology.org/C18-1056/} {Systematic study of long
  tail phenomena in entity linking}.
\newblock In \emph{Proceedings of the 27th International Conference on
  Computational Linguistics}, pages 664--674.

\bibitem[{Le and Titov(2018)}]{le2018improving}
Phong Le and Ivan Titov. 2018.
\newblock \href {https://aclanthology.org/P18-1148/} {Improving entity linking
  by modeling latent relations between mentions}.
\newblock In \emph{Proceedings of the 56th Annual Meeting of the Association
  for Computational Linguistics (Volume 1: Long Papers)}, pages 1595--1604.

\bibitem[{Mendes et~al.(2011)Mendes, Jakob, Garc{\'{\i}}a{-}Silva, and
  Bizer}]{DBLP:conf/i-semantics/MendesJGB11}
Pablo~N. Mendes, Max Jakob, Andr{\'{e}}s Garc{\'{\i}}a{-}Silva, and Christian
  Bizer. 2011.
\newblock \href
  {https://www.google.com/url?sa=t&rct=j&q=&esrc=s&source=web&cd=&cad=rja&uact=8&ved=2ahUKEwig3OnR5cX4AhUFxYUKHeY-BaEQFnoECA0QAQ&url=https%3A%2F%2Fwww.dbpedia-spotlight.org%2Fdocs%2Fspotlight.pdf&usg=AOvVaw2CJ-80y2xV3s-LKc-qiRGF}
  {{DB}pedia {S}potlight: {S}hedding light on the web of documents}.
\newblock In \emph{Proceedings the 7th International Conference on Semantic
  Systems, {I-SEMANTICS} 2011, Graz, Austria, September 7-9, 2011}, pages 1--8.

\bibitem[{Mihalcea and Csomai(2007)}]{10.1145/1321440.1321475}
Rada Mihalcea and Andras Csomai. 2007.
\newblock \href {https://doi.org/10.1145/1321440.1321475} {Wikify! linking
  documents to encyclopedic knowledge}.
\newblock In \emph{Proceedings of the Sixteenth ACM Conference on Conference on
  Information and Knowledge Management}, CIKM '07, page 233–242, New York,
  NY, USA. Association for Computing Machinery.

\bibitem[{Milne and Witten(2008)}]{10.1145/1458082.1458150}
David Milne and Ian~H. Witten. 2008.
\newblock \href {https://doi.org/10.1145/1458082.1458150} {Learning to link
  with wikipedia}.
\newblock In \emph{Proceedings of the 17th ACM Conference on Information and
  Knowledge Management}, CIKM '08, page 509–518, New York, NY, USA.
  Association for Computing Machinery.

\bibitem[{Moro et~al.(2014)Moro, Raganato, and Navigli}]{moro2014entity}
Andrea Moro, Alessandro Raganato, and Roberto Navigli. 2014.
\newblock \href {https://aclanthology.org/Q14-1019/} {Entity linking meets word
  sense disambiguation: a unified approach}.
\newblock \emph{Transactions of the Association for Computational Linguistics},
  2:231--244.

\bibitem[{Piccinno and Ferragina(2014)}]{piccinno2014tagme}
Francesco Piccinno and Paolo Ferragina. 2014.
\newblock \href {https://dl.acm.org/doi/abs/10.1145/2633211.2634350} {From
  {TagME} to {WAT:} a new entity annotator}.
\newblock In \emph{ERD'14, Proceedings of the First {ACM} International
  Workshop on Entity Recognition {\&} Disambiguation, July 11, 2014, Gold
  Coast, Queensland, Australia}, pages 55--62.

\bibitem[{Provatorova et~al.(2021)Provatorova, Bhargav, Vakulenko, and
  Kanoulas}]{provatorova2021robustness}
Vera Provatorova, Samarth Bhargav, Svitlana Vakulenko, and Evangelos Kanoulas.
  2021.
\newblock \href {https://aclanthology.org/2021.emnlp-main.820/} {Robustness
  evaluation of entity disambiguation using prior probes: the case of entity
  overshadowing}.
\newblock In \emph{Proceedings of the 2021 Conference on Empirical Methods in
  Natural Language Processing}, pages 10501--10510.

\bibitem[{R{\"o}der et~al.(2018)R{\"o}der, Usbeck, and
  Ngonga~Ngomo}]{roder2018gerbil}
Michael R{\"o}der, Ricardo Usbeck, and Axel-Cyrille Ngonga~Ngomo. 2018.
\newblock \href {https://dl.acm.org/doi/abs/10.3233/SW-170286}
  {{GERBIL}--benchmarking named entity recognition and linking consistently}.
\newblock \emph{Semantic Web}, 9(5):605--625.

\bibitem[{Tedeschi et~al.(2021)Tedeschi, Conia, Cecconi, and
  Navigli}]{tedeschi2021named}
Simone Tedeschi, Simone Conia, Francesco Cecconi, and Roberto Navigli. 2021.
\newblock \href {https://doi.org/10.18653/v1/2021.findings-emnlp.220} {{N}amed
  {E}ntity {R}ecognition for {E}ntity {L}inking: {W}hat works and what{'}s
  next}.
\newblock In \emph{Findings of the Association for Computational Linguistics:
  EMNLP 2021}, pages 2584--2596, Punta Cana, Dominican Republic. Association
  for Computational Linguistics.

\bibitem[{Usbeck et~al.(2014)Usbeck, Ngomo, R{\"o}der, Gerber, Coelho, Auer,
  and Both}]{usbeck2014agdistis}
Ricardo Usbeck, Axel-Cyrille~Ngonga Ngomo, Michael R{\"o}der, Daniel Gerber,
  Sandro~Athaide Coelho, S{\"o}ren Auer, and Andreas Both. 2014.
\newblock \href
  {https://www.google.com/url?sa=t&rct=j&q=&esrc=s&source=web&cd=&cad=rja&uact=8&ved=2ahUKEwi-wuOi58X4AhXB4YUKHQuzDZ8QFnoECAIQAQ&url=https%3A%2F%2Fsvn.aksw.org%2Fpapers%2F2014%2FECAI_AGDISTIS%2FECAI_short_accepted%2Fpublic.pdf&usg=AOvVaw0PdQTTd_Z9bKnonFoHqlcp}
  {{AGDISTIS}~-~{A}gnostic disambiguation of named entities using linked open
  data.}
\newblock In \emph{ECAI}, volume 2014, pages 1113--1114. Citeseer.

\bibitem[{Wu et~al.(2020)Wu, Petroni, Josifoski, Riedel, and
  Zettlemoyer}]{wu2020scalable}
Ledell Wu, Fabio Petroni, Martin Josifoski, Sebastian Riedel, and Luke
  Zettlemoyer. 2020.
\newblock \href {https://arxiv.org/abs/1911.03814} {Scalable zero-shot entity
  linking with dense entity retrieval}.
\newblock In \emph{Proceedings of the 2020 Conference on Empirical Methods in
  Natural Language Processing (EMNLP)}, pages 6397--6407.

\bibitem[{Yadav and Bethard(2018)}]{yadav-bethard-2018-survey}
Vikas Yadav and Steven Bethard. 2018.
\newblock \href {https://aclanthology.org/C18-1182} {A survey on recent
  advances in named entity recognition from deep learning models}.
\newblock In \emph{Proceedings of the 27th International Conference on
  Computational Linguistics}, pages 2145--2158, Santa Fe, New Mexico, USA.
  Association for Computational Linguistics.

\bibitem[{Yamada et~al.(2016)Yamada, Shindo, Takeda, and
  Takefuji}]{yamada2016joint}
Ikuya Yamada, Hiroyuki Shindo, Hideaki Takeda, and Yoshiyasu Takefuji. 2016.
\newblock \href {https://aclanthology.org/K16-1025/} {Joint learning of the
  embedding of words and entities for named entity disambiguation}.
\newblock In \emph{Proceedings of The 20th SIGNLL Conference on Computational
  Natural Language Learning}, pages 250--259.

\end{thebibliography}
